%% file: RebuttalTemplate.tex
\documentclass[10pt,twocolumn,letterpaper]{article}

\usepackage[pagenumbers]{cvpr} 

\usepackage{graphicx}
\usepackage{amsmath}
\usepackage{amssymb}
\usepackage{booktabs}
\usepackage{caption}
\usepackage{subcaption}
\usepackage{titling}
\usepackage{tabularx}

\usepackage[pagebackref,breaklinks,colorlinks]{hyperref}

\usepackage[capitalize]{cleveref}
\crefname{section}{Sec.}{Secs.}
\Crefname{section}{Section}{Sections}
\Crefname{table}{Table}{Tables}
\crefname{table}{Tab.}{Tabs.}

\begin{document}

\title{Joint Coordinate Regression and Association For Multi-Person Pose Estimation, A Pure Neural Network Approach}

\author{Dongyang Yu Email: \href{mailto:yudongyang2022@gmail.com}{yudongyang2022@gmail.com}Yunshi Xie Email: \href{mailto:Yunshix2@illinois.edu}{Yunshix2@illinois.edu}\\Wangpeng An Email: \href{mailto:anwangpeng@gmail.com}{anwangpeng@gmail.com}\\Li Zhang Email: \href{mailto:zhang_li@bjfu.edu.cn}{zhang-li@bjfu.edu.cn}\\Yufeng Yao Email: \href{mailto:yaoyufeng@bjfu.edu.cn}{yaoyufeng@bjfu.edu.cn}}

\maketitle
\input{abstract.tex}

\input{introduction.tex}

\input{related.tex}

\input{method.tex}

\input{experiment.tex}

\input{conclusion.tex}

{\small
\bibliographystyle{ieee_fullname}
\bibliography{egbib}
}

\end{document}

%% file: abstract.tex
\begin{abstract}
We introduce a novel one-stage end-to-end multi-person 2D pose estimation algorithm, known as Joint Coordinate Regression and Association (JCRA), that produces human pose joints and associations without requiring any post-processing. The proposed algorithm is fast, accurate, effective, and simple. The one-stage end-to-end network architecture significantly improves the inference speed of JCRA. Meanwhile, we devised a symmetric network structure for both the encoder and decoder, which ensures high accuracy in identifying keypoints. It follows an architecture that directly outputs part positions via a transformer network, resulting in a significant improvement in performance. Extensive experiments on the MS COCO and CrowdPose benchmarks demonstrate that JCRA outperforms state-of-the-art approaches in both accuracy and efficiency. Moreover, JCRA demonstrates 69.2 mAP and is 78\% faster at inference acceleration than previous state-of-the-art bottom-up algorithms. The code for this algorithm will be publicly available.
\end{abstract}

%% file: introduction.tex
\section{Introduction}
One of the fundamental and compelling computer vision tasks is human pose estimation (HPE),  
and this task has attracted intense attention in recent years. With the increase of computing power, algorithms with higher accuracy and faster computing speed are emerging.
Compared with single-person pose estimation, the multi-person pose estimation(MHPE) is more complex as it aims to detect all the instances and identify joints of each person. The importance of MHPE mainly comes from a large set of application~\cite{YongDu2015HierarchicalRN,MykhayloAndriluka2017PoseTrackAB,DevaRamanan2005StrikeAP} based on it, i.e., human behaviors understanding, motion capture, violence detection, pedestrian tracking, crowd riot scene identification, human-computer interaction, and autonomous driving, et. al

In order to solve MHPE task, there are two mainstream methods exists with two-stage frameworks: top-down~\cite{GedasBertasius2019LearningTP,HaoShuFang2016RMPERM,ZhenguangLiu2021DeepDC, AlejandroNewell2016StackedHN, KeSun2019DeepHR, ShihEnWei2016ConvolutionalPM} and bottom-up~\cite{ZheCao2022RealtimeM2,ZigangGeng2021BottomUpHP, ShengJin2020DifferentiableHG, SvenKreiss2019PifPafCF,FangyunWei2020PointSetAF} approaches.
the top-down paradigm takes two stages, the first is to detect human bounding boxes and then perform single person pose estimation for each bounding box. For the human detection stage, top-down algorithms are more accurate as the algorithm transformed the problem into a single-person-pose estimation problem by utilizing the object detection algorithm to find each person. However, 
 the computational cost is large and expensive, As the number of human instances in images increases. The pose estimation accuracy depends heavily on the performance of the detection network. As a result, most real-world applications are based on bottom-up algorithms.   

The other branch is bottom-up, which first locates identity-free keypoints and then group them into different person.The grouping process is usually heuristic,hand-crafted and tedious, with lots of tricks to adjust hyper-parameters and this work may cost a lot of manpower.

In order to overcome the shortcomings of the above two methods, a new method has recently been proposed, which directly estimates multi-person poses from the input image in one single stage~\cite{stoffl2021end, WeianMao2022FCPoseFC,XuechengNie2019SingleStageMP, DahuShi2021InsPoseIN, ZhiTian2019DirectPoseDE,FangyunWei2020PointSetAF, petr}. FCpose~\cite{WeianMao2022FCPoseFC} and InsPose~\cite{DahuShi2021InsPoseIN} propose a fully convolutional multi-person pose estimation framework which is efficient with the dynamic instance-aware convolutions. SPM~\cite{XuechengNie2019SingleStageMP} proposes a structured pose representation that makes person instance and body joint position representation to be a whole unit.

\begin{figure*}
\begin{subfigure}{\textwidth}
  \centering
\includegraphics[height=4in,width=\linewidth]{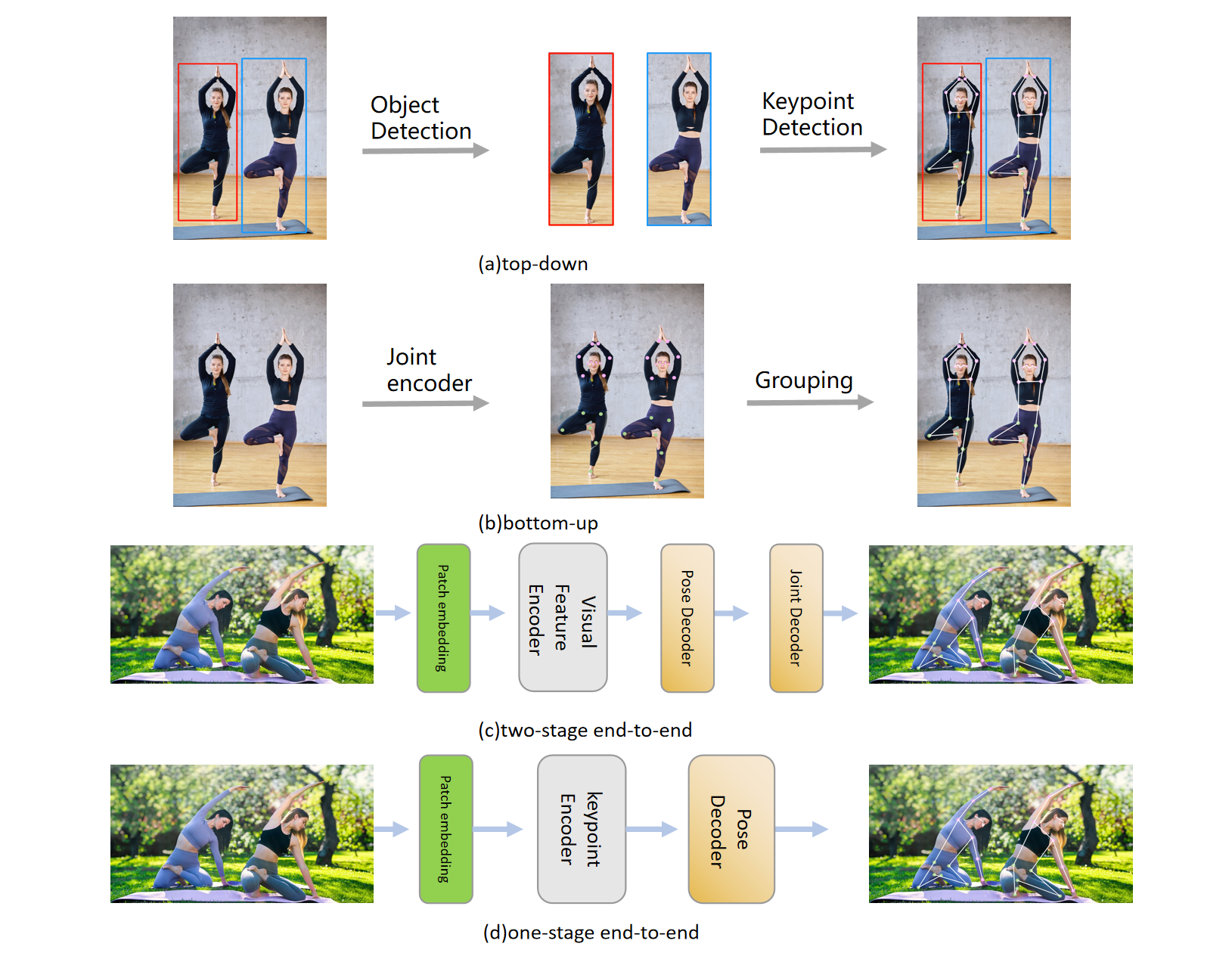}  

  \label{fig:sub-first}
\end{subfigure}

\caption{Overview of (a) top-down, (b) bottom-up, (c) two-stage end-to-end methods and (d) one sate end-to-end methods. The last row show the overview of our Joint Coordinate Regression and Association (JCRA) algorithm. JCRA is a one-stage end-to-end method.}
\label{fig:overview}
\end{figure*}

Inspired by recent work on end-to-end trainable object detection using Transformer, POET~\cite{stoffl2021end} proposed an end-to-end trainable approach for MHPE, Combining a convolutional neural network with a transformer encoder-decoder architecture, POET formulate multi-instance pose estimation from images as a direct set prediction problem. 
PETR~\cite{petr} describes an end-to-end paradigm with one-stage Transformer framework, in PETR multi-person pose queries are learned to directly reason a set of full-body poses.First pose decoder predicts multi full-body poses in parallel, Then a joint decoder is utilized to further refine the poses by exploring the kinematic relations between body joints. This end-to-end framework removes some post-processing like RoI cropping, NMS, and grouping.

The aforementioned end-to-end methods achieve a good trade-off between accuracy and efficiency despite they don't need the extra RoI (Region of Interest) cropping and keypoints grouping post-progressing. However, heatmap~\cite{XuechengNie2019SingleStageMP,zhou2019objects} or score map~\cite{WeianMao2022FCPoseFC, DahuShi2021InsPoseIN} or NMS (Non-Maximum Supression)~\cite{WeianMao2022FCPoseFC,DahuShi2021InsPoseIN,FangyunWei2020PointSetAF} and decoder data fusion~\cite{petr} post-progressing are need although, which are not really end-to-end optimized MHPE.
To simplify the pipeline further, i.e. remove the postprocessing part and concentrate all operations on neural networks, We propose a novel JCRA for human pose estimation, which deals with the aforementioned problem effectively. The JCRA makes full use of the transformer's advantage allowing for estimating keypoint directly and accurately without heatmaps. The final network realizes a notable performance enhancement compared to start-of-the-art results on the COCO val2017 dataset~\cite{cocodata}.
The main contributions of this paper are:
\begin{itemize}
    \item We propose a real end-to-end learning framework for MHPE named JCRA. The proposed JCRA method directly predicts full-body pose without any post-processing.
    \item JCRA surpasses all end-to-end methods and has obtained a very close score in large-scale MHPE tasks compared with top-down methods.
    \item  JCRA predicts the multi-person pose of each individual in the image and video very fast with the highest accuracy. Compared with the current end-to-end State-Of-The-Art algorithm PETR~\cite{petr}, JCRA's inference speed is almost increased by 2 times. 
\end{itemize}

%% file: related.tex
\section{Related Works}
The pose estimation has developed from CNNs~\cite{BinXiao2018SimpleBF} to vision transformer networks. Early works applied transformers as a better decoder~\cite{KeLi2021PoseRW,YanjieLi2021TokenPoseLK,SenYang2020TransPoseKL}, e.g., TransPose~\cite{SenYang2020TransPoseKL} directly processes the features extracted by CNNs to model the global relationship. TokenPose~\cite{YanjieLi2021TokenPoseLK} proposes token-based representations by introducing extra tokens to estimate the locations of occluded keypoints and model the relationship among different keypoints. To get rid of the CNNs for feature extraction, HRFormer~\cite{YuhuiYuan2021HRFormerHT} is proposed to use transformers to extract high-resolution features directly. A delicate parallel transformer module is proposed to fuse multi-resolution features in HRFormer gradually. These transformer-based pose estimation methods obtain superior performance on popular keypoint estimation benchmarks. In Fig. \ref{fig:overview}, The pose estimation work branched out bottom-up, top-down and recent two-stage end-to-end methods.

\subsection{Top-down MHPE methods}

Two stages are involved in a top-down pose estimation algorithm: detecting human bounding boxes and then estimating a single person's pose for each bounding box. The typical top-down algorithm includes Hourglass~\cite{AlejandroNewell2016StackedHN}, RMPE~\cite{HaoShuFang2017RMPERM}, CPN~\cite{YilunChen2017CascadedPN}, SimpleBaseline~\cite{BinXiao2018SimpleBF}, HRNet~\cite{KeSun2019DeepHR}, which detect the keypoints of a single person within a person bounding box. The person bounding boxes are usually generated by an object detector~\cite{faster}. Mask R-CNN~\cite{mask} directly adds a keypoint detection branch on Faster R-CNN~\cite{faster} and reuses features after ROIAlign. All the aforementioned algorithms utilized heatmaps to encode and decode keypoints, an following those algorithms, Simcc~\cite{coordeccv} applied direct regression of the coordinates of the key points by the negative log-likelihood loss function. In short, Top-down methods are accurate but highly dependent on the detector's performance, and incorporating a personal detector would cost more running time.

\begin{figure*}
\begin{subfigure}{\textwidth}
  \centering
\includegraphics[height=3in,width=\linewidth]{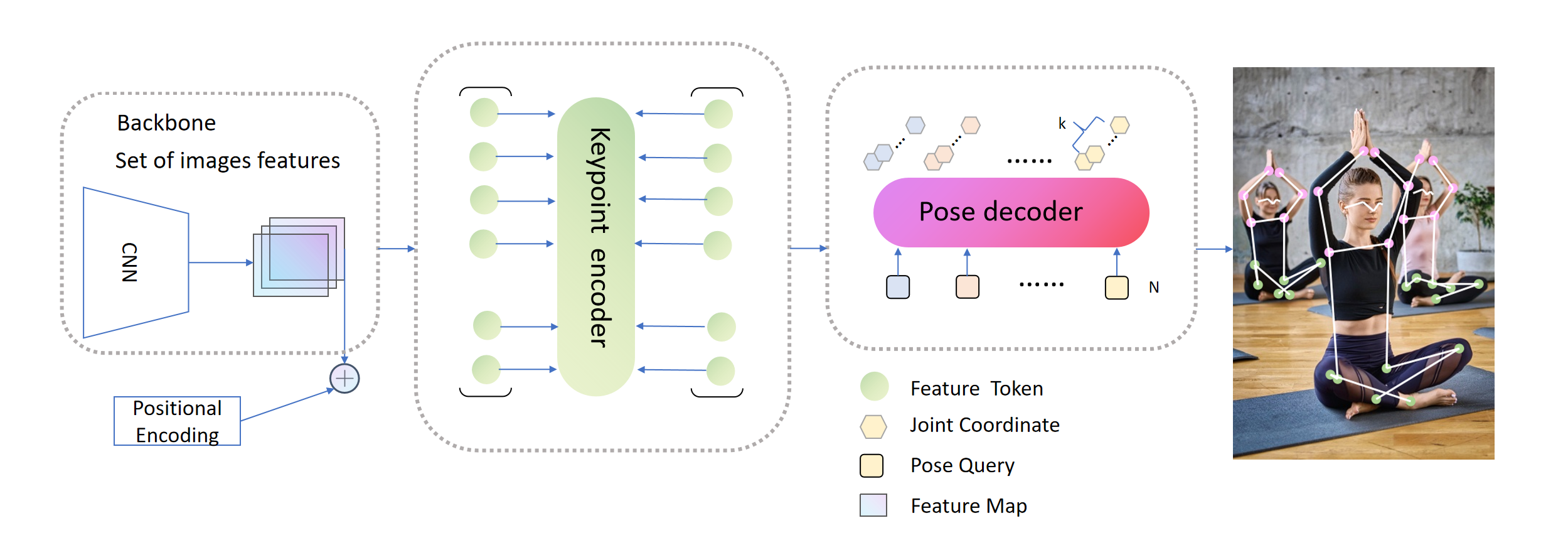}  

  \label{fig:sub-first}
\end{subfigure}

\caption{Overview of Joint Coordinate Regression and Association (JCRA) algorithm}
\label{fig:overview2}
\end{figure*}

\subsection{Bottom-up MHPE alogirhtms}

The pose estimation usually contains key points detection and associations, most works performed the procedure independently until Openpose~\cite{realtime}, Which learning the fittest association of joints,  regardless of the number of people in the image, which is an effective non-parametric representation. In addition, this method is bottom-up and consists of only one stage so that the pose can be served in real time. Later, associative embedding~\cite{assocative} method were proposed, where the network outputs the detection and grouping simultaneously. Those papers remarks on the transition from two-stage pose estimation to a single stage. \par 

The stacked hourglass network~\cite{sh} plays a key role, where the features of different scales can be combined and consolidated to get the relevant spatial connections associated with the body part. The symmetric architecture results in an improvement in both performance and efficiency. Our network design is enlightened by the stacked hourglass network to some degree.

\subsection{End-to-end MHPE methods}

The top-down and bottom-up methods are two polar of the solutions, in order to balance them, the single-stage methods~\cite{WeianMao2022FCPoseFC,XuechengNie2019SingleStageMP,FangyunWei2020PointSetAF,DahuShi2021InsPoseIN,stoffl2021end,petr} are proposed, which densely regress a set of pose candidates over spatial locations, where each candidate consists of the keypoint positions that are from the same person. POET~\cite{stoffl2021end} proposed an end-to-end trainable approach for MHPE, which combines a convolutional neural network with a transformer encoder-decoder architecture, POET formulates multi-instance pose estimation from images as a direct set prediction problem. 
PETR~\cite{petr} describes an end-to-end paradigm with one-stage Transformer framework, in PETR multi-person pose queries are learned to directly reason a set of full-body poses. Followed by a  pose decoder that predicts multi full-body poses in parallel, Then a joint decoder is utilized to further refine the poses by exploring the kinematic relations between body joints. This end-to-end framework removes some post-processing like part RoI cropping, pose-NMS, and grouping. 

The transformer~\cite{detr} based approaches are the new trend as researchers are exploring attention mechanisms for human pose estimation.
The transformer is an architecture of encoder-decoder, the encoder-decoder framework in general refers to a situation in which one process represents, or "encodes" input data into one vector, and then another process "decodes" that vector into the desired output. In order for the decoder to be able to use all of the hidden states, this matrix needs to be condensed into a vector of consistent size - we cannot pass a hidden state into the decoder whose size varies based on the number of elements in the source sequence. Consequently, attention was born.
\begin{equation}
\text{Attention}(Q', K', V') = \text{softmax}(\frac{Q'K'^T}{\sqrt{d_k}})V'
\label{eq:attention}
\end{equation}
Where $\in \mathbb R^{m \times d_{model}}$, $K \in \mathbb R^{n \times d_{model}}$ and $V \in \mathbb R^{n \times d_{model}}$.

Motivated by ~\cite{soft_attention}
Transformers have been applied to the pose estimation task with some success. TransPose~\cite{transpose} and HRFormer~\cite{hrformer} enhance the backbone via applying the Transformer encoder to the backbone; TokenPose~\cite{tokenpose} designs the pose estimation network in a ViT-style fashion by splitting image into patches and applying class tokens, which makes the pose estimation more explainable. These methods are all heatmap-based and use a heavy transformer encoder to improve the model capacity. In contrast, Poseur~\cite{poseur} is a regression-based method with a lightweight transformer decoder. Thus, Poseur is more computationally efficient while can still achieve high performance. PRTR~\cite{prtr} leverages the encoder-decoder structure in transformers to perform pose regression. PRTR is based on DETR~\cite{detr}, i.e., it uses Hungarian matching strategy to find a bipartite matching between non-class-specific queries and  ground-truth joints.

Early works tend to treat the transformer as a better decoder~\cite{prtr, tokenpose, transpose}, e.g., TransPose~\cite{transpose} directly
processes the features extracted by CNNs to model the global relationship. TokenPose ~\cite{tokenpose} proposes
token-based representations by introducing extra tokens to estimate the locations of occluded keypoints and model the relationship among different keypoints. To get rid of the CNNs for feature
extraction, HRFormer~\cite{hrformer} is proposed to use transformers to extract high-resolution features directly.
A delicate parallel transformer module is proposed to fuse multi-resolution features in HRFormer
gradually. These transformer-based pose estimation methods obtain superior performance on popular
keypoint estimation benchmarks. However, they either need CNNs for feature extraction or require
careful designs of the transformer structures. ViTPose~\cite{vitpose} explored the plain vision transformers for the pose estimation tasks. 
PETR~\cite{petr} adapted DETR~\cite{detr} by directly output person key points, bounding box, and person probability.
logocap~\cite{logocap} learning the local-global contextual adaptation for bottom-up human pose estimation.

%% file: method.tex
\section{Algorithm}

\subsection{The Motivation}
Multi-person pose estimation has gained great progress in recent years~\cite {cpm,assocative, accurate, realtime, articulated,eichner2009better, petr, coordeccv}, which can be categorized as keypoints detection and association briefly. 
Complex transformations are applied in either top-down or bottom-top algorithms, i.e. the representation and decode of the heatmaps in order to estimate keypoint location accurately, the representations of the keypoint associations like part affinity field~\cite{realtime} and associative embedding~\cite{assocative}. Most of the complexity is comes from post-processing, which solved by non-trivial methods. Simplifying the whole pipeline would be valuable to the community.
\subsection{Joint Coordinate Regression and Association (JCRA)}
The proposed Joint Coordinate Regression and Association can output keypoint coordinates directly without heatmaps, and the association of those keypoints by the query of the transformer. The proposed algorithm is one-stage end-to-end, all operations are performed by neural networks, we illustrated the architecture of our method in Fig. \ref{fig:overview2}. 
\subsection{Backbone}
A Resnet50~\cite{resnet} is applied to extract features of the image, followed by feature pyramid networks to extract
multi-level feature maps. Flattened image features are fed into the keypoint encoder and refined. In order to demonstrate the effectiveness and robustness of JCRA, we used the ResNet50 backbone on the Coco dataset, and the Swin-L backbone on the CrowdPose dataset.
\subsection{Keypoint Encoder}
The keypoint encoder is applied to refine the multi-scale feature maps. Before passing flattened image features to the keypoint encoder, positional encoding is added. A feed-forward network (FFN) and a multi-scale deformable attention~\cite{deformatten} module are present in each encoder layer. In keypoint encoder, $6$ deformable encoder layers are stacked in sequence.
\subsection{Pose Decoder}
In a similar way to the keypoint encoder, we use the deformable attention module to build the keypoint decoder. The keypoint decoder predicts $300$ full-body keypoints in parallel based on the $300$ keypoint queries and the refined multi-scale feature tokens. The $300$ keypoint queries are transformed into an output embedding by the decoder. They are then independently decoded into key points and scores by
a feed-forward network (FFN), resulting in $300$ final
predictions. The shape of the final tensor is $300$x$17$x$3$. The $17$ is the number of keypoints in COCO dataset, while $3$ is (x,y,confidence) of each keypoint, (x,y) is the coordinate of keypoint.

\begin{table}[]
\scriptsize
\centering
\begin{tabular}{l|c|c|c|c}
\hline
\textbf{Frameworks} & \textbf{RoI-free} & \textbf{Grouping-free} & \textbf{NMS-free} & \textbf{Refine-free} \\ \hline
Top-down &    &  $\checkmark$  &  \\ 
Bottom-up & $\checkmark$ &    &    \\ \hline
Non end-to-end & $\checkmark$ & $\checkmark$ &   \\ 
Two-stage end-to-end &  $\checkmark$  &  $\checkmark$  & $\checkmark$ \\ 
One-stage end-to-end & $\checkmark$ & $\checkmark$ & $\checkmark$ & $\checkmark$ \\ 
\hline
\end{tabular}
\caption{Comparison of pose estimation frameworks}
\label{tbk:frameworks}
\end{table}

\subsection{Loss Functions}
In accordance with \cite{detr}, we employ a set-based Hungarian loss, ensuring a distinct prediction for each ground-truth pose. Our pose decoder's classification head utilizes the same classification loss function ($L_{c}$) as in \cite{zhu2021deformable}. $L_{c}$ is the focal loss function \cite{lin2017focal} for the classifcation branch. Additionally, we incorporate $L_1$ loss ($L_{reg}$) and OKS loss ($L_{oks}$) for pose regression heads in our pose decoder, respectively. Analogous to \cite{DahuShi2021InsPoseIN}, we employ auxiliary heatmap regression training for rapid convergence. We employ a deformable transformer encoder to generate heatmap predictions. Subsequently, we calculate a variant of focal loss \cite{law2018cornernet} between the predicted and ground-truth heatmaps, denoted as $L_{hm}$. 
The comprehensive loss function of our model can be expressed as:

\begin{equation}
\text L = L_{c} + + \lambda_{1} L_{hm} + \lambda_{2} L_{reg} + \lambda_{3} L_{oks}  
\label{eq:attention}
\end{equation}

The loss weights are denoted by $\lambda_{1}$, $\lambda_{2}$, and $\lambda_{3}$.

\subsection{One-stage Vs Two-stage End-to-end Framework}
In Table~\ref{tbk:frameworks}, we compare JCRA with other frameworks for pose estimation. JCRA represents a comprehensive, one-stage end-to-end framework. The one-stage end-to-end framework for human posture estimation offers advantages such as high efficiency, enhanced accuracy, and robust performance. By integrating multiple processing stages, this approach minimizes computational time and resource demands, enabling real-time processing. The one-stage end-to-end framework has exhibited considerable speed advantages in both model training and model inference.

The two-stage end-to-end framework, similar to PETR\cite{petr}, comprises two decoders: the primary decoder and the refinement decoder. The refinement decoder serves to enhance the detection accuracy of key points. As observed in Table~\ref{tbk:results}, PETR\cite{petr} demonstrates a superior keypoint detection capability for challenging samples compared to JCRA. Nonetheless, for large and medium-sized targets, JCRA attains greater accuracy. Analogous occurrences were also observed in the COCO validation and COCO test datasets. We concluded that the one-stage end-to-end framework excels in detecting keypoints for large and medium-sized targets. Conversely, the two-stage end-to-end framework, incorporating a refinement decoder, exhibits an improvement in keypoint detection capabilities for difficult targets.

\begin{figure*}
\begin{subfigure}{0.33\textwidth}
  \centering
  \includegraphics[height=1.7in,width=\linewidth]{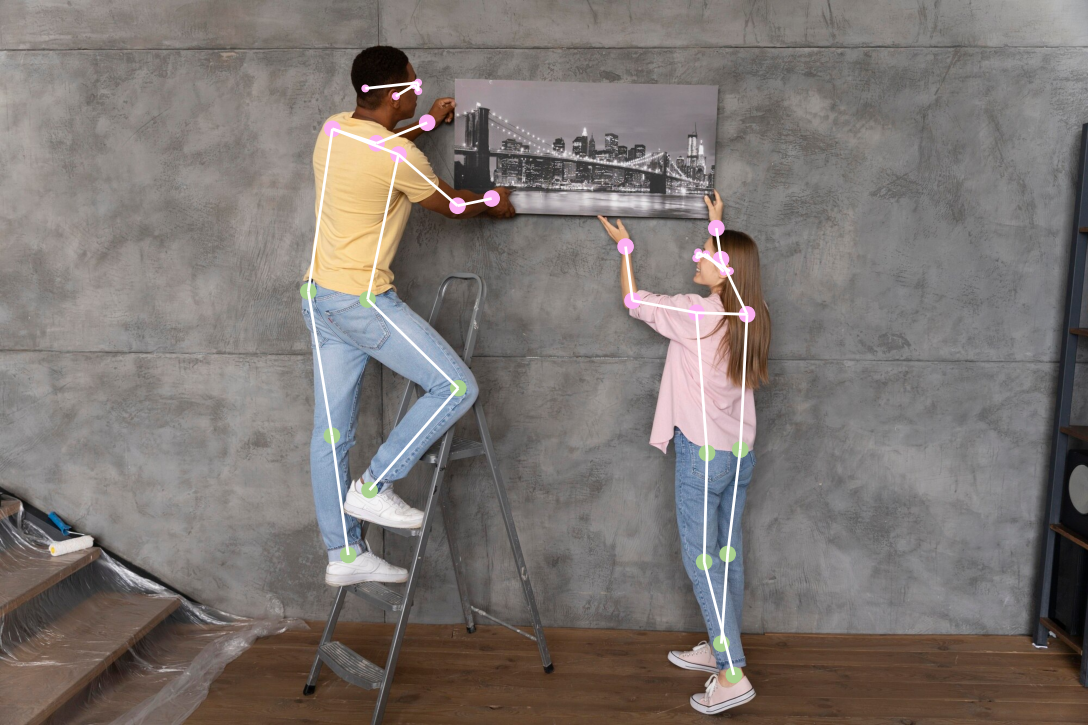}  
  \label{fig:sub-first}
\end{subfigure}
\begin{subfigure}{.33\textwidth}
  \centering
  \includegraphics[height=1.7in,width=\linewidth]{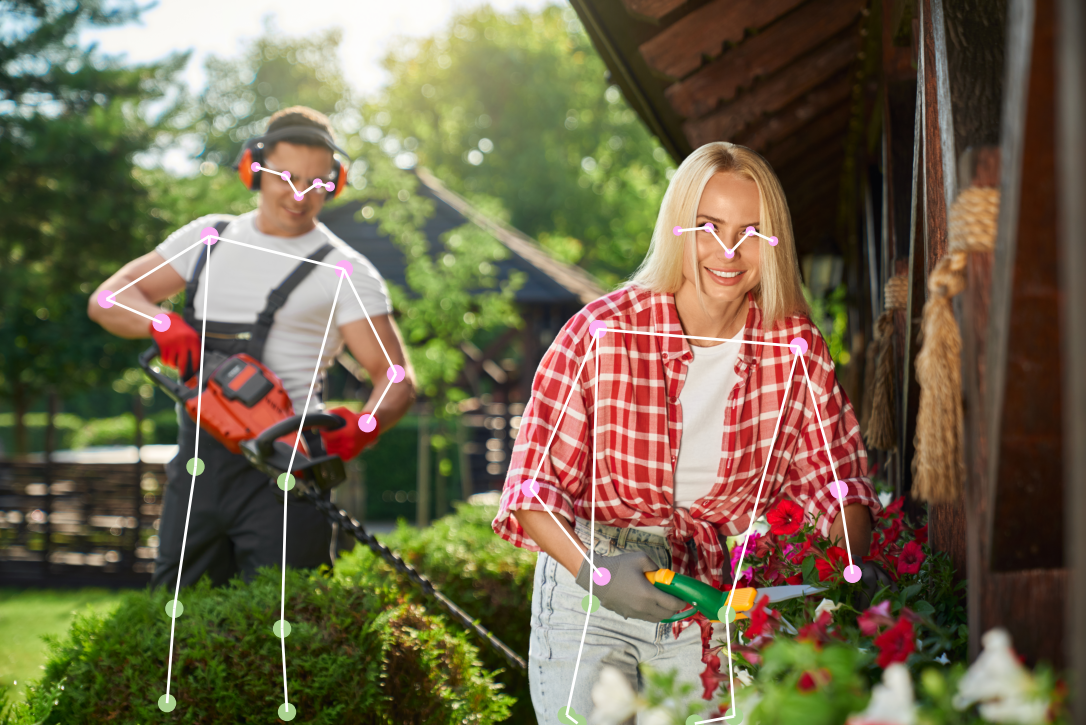}  
  \label{fig:sub-second}
\end{subfigure}
\begin{subfigure}{.33\textwidth}
  \centering
  \includegraphics[height=1.7in,width=\linewidth]{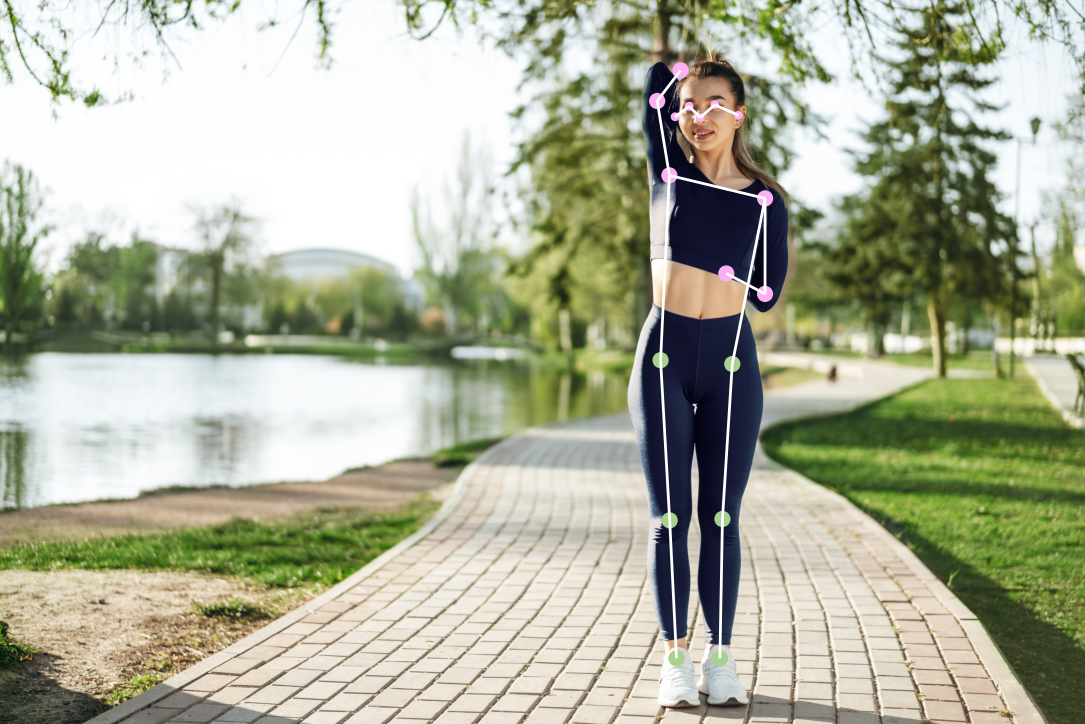}  
  \label{fig:sub-second}
\end{subfigure}
\newline
\begin{subfigure}{.33\textwidth}
  \centering
  \includegraphics[height=1.7in,width=\linewidth]{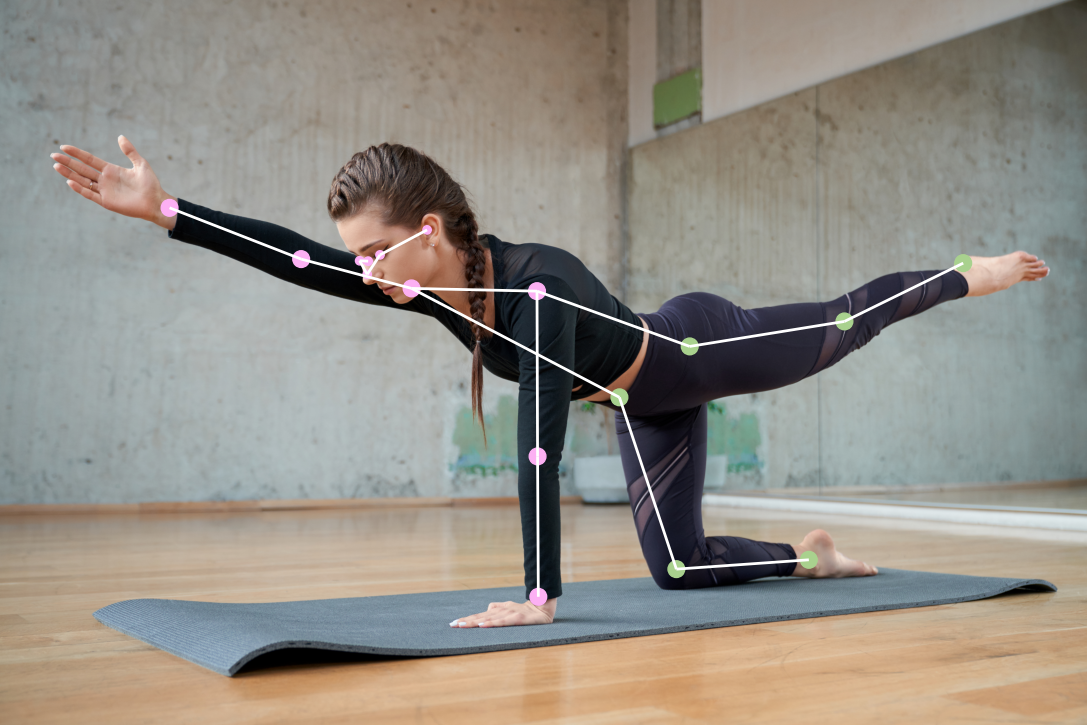}  
  \label{fig:sub-third}
\end{subfigure}
\begin{subfigure}{.33\textwidth}
  \centering
  \includegraphics[height=1.7in,width=\linewidth]{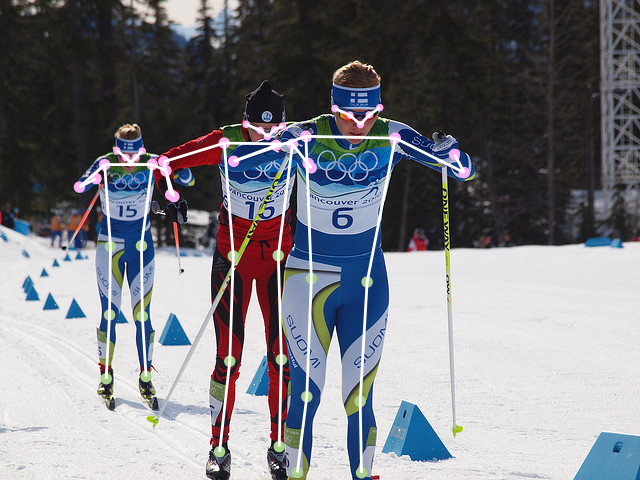}  
  \label{fig:sub-fourth}
\end{subfigure}
\begin{subfigure}{.33\textwidth}
  \centering
  \includegraphics[height=1.7in,width=\linewidth]{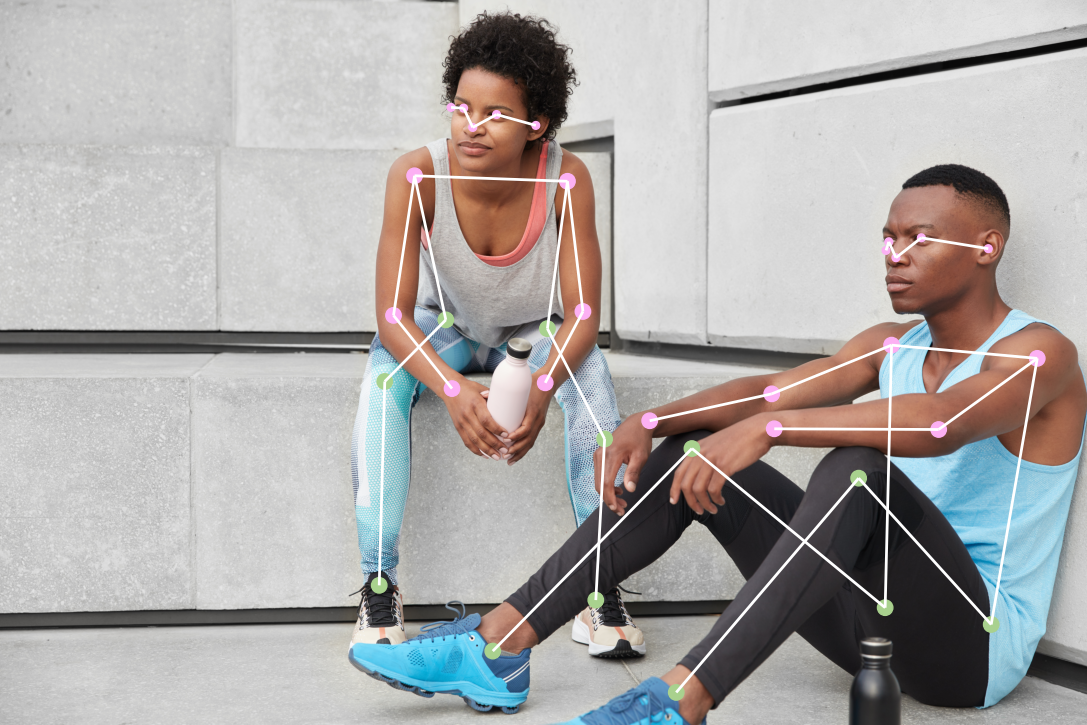}  
  \label{fig:sub-fourth}
\end{subfigure}
\caption{The visualization results of the JCRA. The first row and the second row show the visualization results on COCO dataset, respectively. A wide range of poses can be handled by JCRA, including viewpoint change, occlusion, and crowded settings. }
\label{fig:fig}
\end{figure*}

\subsection{Symmetrical Network Architecture}

Symmetry in the design, characterized by an equal number of layers in both the encoder and decoder, guarantees that each level of abstraction in the encoder corresponds to a level in the decoder, facilitating the translation of abstractions back into more concrete forms.
   
\begin{figure}
\begin{subfigure}{0.48\textwidth}
  \centering
\includegraphics[height=2.2in,width=\linewidth]{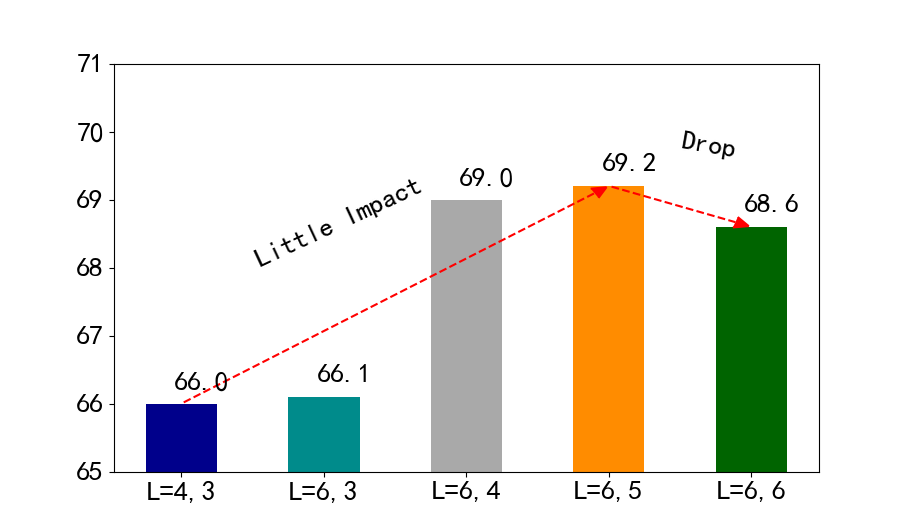}  

  \label{fig:sub-layers}
\end{subfigure}

\caption{L represents the encoder and decoder layers. L =4,3 means that the number of layers of the keypoint encoder is 4, and the number of layers of pose decoder is 3. When L = 6,5, we got the highest score  69.2 mAP on COCO  val2017 dataset.}
\label{fig:layers}
\end{figure}

The JCRA employs an approximately symmetric network architecture, consisting of six encoder layers and five decoder layers. The advantages of a symmetric encoder and decoder include: A reduction in information loss during the encoding and decoding processes, leading to enhanced reconstruction quality of the output. A more balanced model that effectively captures pertinent features of the input data during both encoding and decoding. The model can acquire more comprehensive and stable representations of the input data that exhibit a high tolerance towards variations and noise. Owing to its symmetric network architecture, the JCRA attains a 69.2 mAP without necessitating keypoint refinement. When the number of layers in the encoder and decoder networks is proximate, higher accuracy can be achieved, as illustrated in Fig. \ref{fig:layers}.

\begin{table*}[h]
\scriptsize
\centering
\begin{tabular}{c|c|c c c|c c|c}
\hline
Method&Backbone&$AP^{kp}$&$AP_{50}^{kp}$&$AP_{75}^{kp}$&$AP_{M}^{kp}$&$AP_{L}^{kp}$&Speed(fps)\\
\hline
&&&Top-down methods&&&\\
\hline
Mask R-CNN~\cite{mask}&ResNet-50&64.2& 86.6&69.7 &58.7 &73.0&8.2\\
Simcc~\cite{coordeccv}&ResNet-50&73.0&89.3&79.7 & \textbf{69.5}&\textbf{79.9}&-\\
Poseur~\cite{poseur}&ResNet-50&75.4&90.5&82.2 & 68.1&78.6&-\\
ViTPose ~\cite{vitpose}&ViTPose-B&\textbf{75.8}&\textbf{90.7} &\textbf{83.2}& 68.7&78.4&-\\
\hline
&&&Bottom-up methods&&&\\
\hline
CMU-Pose~\cite{realtime}&VGG-19&65.3&85.2&71.3 & 62.2&70.7 & 10.1\\
DEKP~\cite{dekp}&HRNet-W32&68.0&86.7&74.5&62.1&77.7 &1.8\\
HrHRNet~\cite{cheng2020higherhrnet}&HRNet-W32&68.5&-&- &\textbf{64.3}&75.3&1.8\\
\hline
&&&Two-stage end-to-end methods&&&\\
\hline
PETR~\cite{petr}&ResNet-50&68.8&87.5&76.3 & 62.7&77.7 &8.2\\
\hline
&&&Fully one-stage end-to-end methods&&&\\
\hline
JCRA(Ours)~&ResNet-50&\textbf{69.2}&\textbf{89.4}&\textbf{76.7} &63.0&\textbf{78.3}&\textbf{14.6}\\
\hline
\end{tabular}
\caption{Keypoint detection AP on COCO val2017 dataset}
\label{tbl:results}
\end{table*}

\begin{table*}[h]
\scriptsize
\centering
\begin{tabular}{c|c|c c c|c c|c}
\hline
Method&Backbone&$AP^{kp}$&$AP_{50}^{kp}$&$AP_{75}^{kp}$&$AP_{M}^{kp}$&$AP_{L}^{kp}$&Speed(fps)\\
\hline
&&&Top-down methods&&&\\
\hline
Mask R-CNN~\cite{mask}&ResNet-50&62.7 &87.0& 68.4& 57.4 &71.1&8.2\\
PRTR~\cite{prtr}&HRNet-W32&72.1 &90.4& 79.6& 68.1 &79.0&-\\
Simcc†~\cite{coordeccv}&ResNet-50&72.7& 91.2 &80.1 &69.2 &79.0&-\\
\hline
&&&Bottom-up methods&&&\\
\hline
CMU-Pose~\cite{realtime}&VGG-19&64.2 &86.2 &70.1 &61.0 &68.8& 10.1\\
DEKP~\cite{dekp}&HRNet-W32&67.3 &87.9 &74.1 &61.5& 76.1&1.8\\

HrHRNet~\cite{cheng2020higherhrnet}&HRNet-W32&66.4 &87.5 &72.8 &61.2& 74.2&1.8\\
\hline
&&&Non end-to-end methods&&&\\
\hline
DirectPose ~\cite{ZhiTian2019DirectPoseDE}&ResNet-50&62.2& 86.4 &68.2 &56.7 &69.8&9.9\\
FCPose ~\cite{WeianMao2022FCPoseFC}&ResNet-50&64.3& 87.3 &71.0& 61.6 &70.5&10.7\\
InsPose ~\cite{DahuShi2021InsPoseIN}&ResNet-50&65.4& 88.9& 71.7& 60.2& 72.7&9.1\\
DirectPose ~\cite{ZhiTian2019DirectPoseDE}&ResNet-101&63.3 &86.7& 69.4 &57.8 &71.2&-\\
FCPose ~\cite{WeianMao2022FCPoseFC}&ResNet-101&65.6& 87.9& 72.6 &62.1& 72.3& 7.8\\
InsPose ~\cite{DahuShi2021InsPoseIN}&ResNet-101&66.3& 89.2& 73.0& 61.2& 73.9& 7.3\\

\hline
&&&Two-stage end-to-end methods&&&\\
\hline
PETR~\cite{petr}&ResNet-50&67.6 &89.8 &75.3 &61.6 &76.0&8.2\\
\hline
&&&Fully one-stage end-to-end methods&&&\\
\hline
JCRA(Ours)~&ResNet-50&\textbf{67.6}&\textbf{90.0}&75.2 &\textbf{61.6}&\textbf{76.1}&\textbf{14.6}\\
\hline
\end{tabular}
\caption{Keypoint detection AP on COCO test-dev dataset. The symbol † is used to indicate the adoption of Gaussian label smoothing in our experiments.}
\label{tbm:results}
\end{table*}

%% file: experiment.tex
\section{Experimental Results}

\subsection{COCO Keypoint Detection}
We perform a thorough comparison of the improvement of JCRA to state-of-the-art algorithms in human pose estimation, and the experiment datasets used for evaluation are COCO~\cite{cocodata}, which consists of about $200K$ images and contain around $250K$ people with annotated key points, the dataset contains diverse scenarios such as scale variation, different camera views, and occlusion in the image. We employed the val2017 dataset, which comprises 5,000 images, to validate our ablation experiments. In addition, we compared our method to other state-of-the-art approaches on the test-dev set that contains 20,000 images.

\textbf{Evaluation metrics.} The standard evaluation metric
is based on Object keypoint Similarity (OKS). We report
standard average precision and recall scores\footnote{http://cocodataset.org/\#keypoints-eval}: $AP_{50}^{kp}$ ($AP^{kp}$
at OKS = 0.50), $AP_{75}^{kp}$, $AP^{kp}$ (mean of AP scores from
OKS = 0.50 to OKS = 0.95 with the increment as 0.05),
$AP_{M}^{kp}$ for persons of medium sizes and $AP_{L}^{kp}$ for persons of
large sizes.

\subsection{Results on the COCO val2017 dataset}
We compare JCRA to the state-of-the-art methods in pose estimation in Table~\ref{tbl:results}. Our method achieves best performance among all bottom-up algorithms, and comparable performance with top-down human pose estimation algorithms, The compared algorithms include CMU-Pose~\cite{realtime} and Mask R-CNN~\cite{mask}, PETR~\cite{petr} and Simcc~\cite{coordeccv}, VitPose ~\cite{vitpose}, the state-of-the-art top-down pose estimation algorithm, respectively. Visualization outputs by JCRA are presented in Fig. \ref{fig:fig}. The network makes competitive results even in the condition that large overlaps or congestion exist among multiple people.\par 
Table~\ref{tbl:results} shows that our result (69.2 AP) is $0.58\%$ higher than the latest PETR~\cite{petr} for two-stage end-to-end methods. Our method is much simple and more effective. Our algorithm even suppressed top-down methods regarding the $AP_{L}^{kp}$ metric.

\subsection{Results on the COCO test-dev}
Initially, we compared our Joint Coordinate Regression and Association (JCRA) with other state-of-the-art methods, as presented in Table~\ref{tbm:results}. Utilizing the same backbone network for feature extraction, our JCRA surpasses all existing bottom-up approaches and end-to-end methods. Remarkably, without further enhancements, our model achieved an impressive AP score of 67.6 using ResNet-50.

\begin{figure}
\begin{subfigure}{0.45\textwidth}
  \centering
\includegraphics[height=2.2in,width=\linewidth]{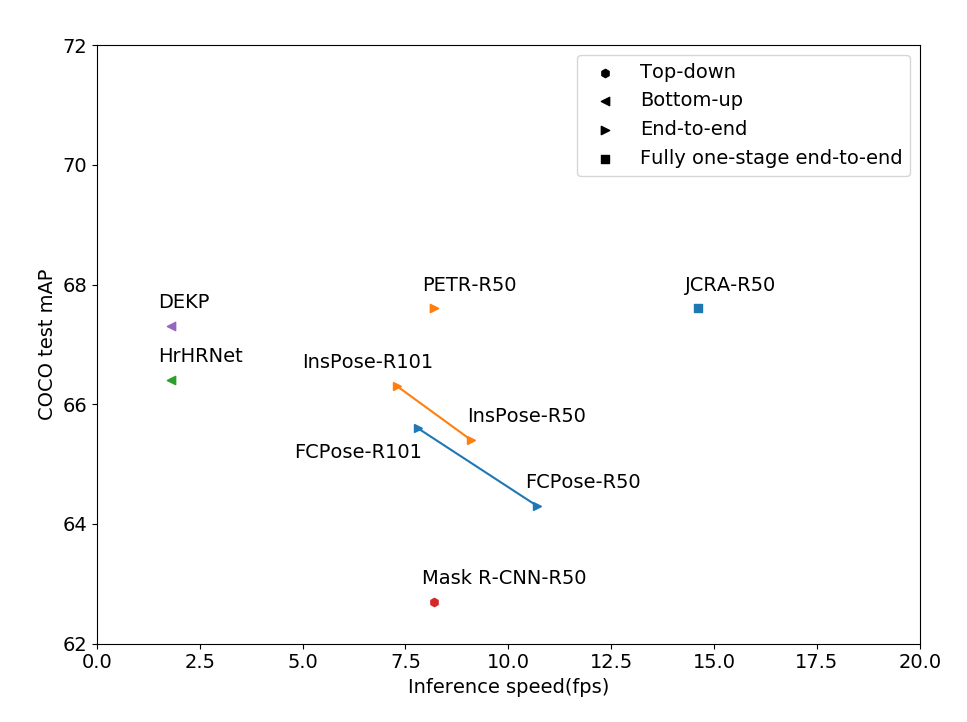}  

  \label{fig:sub-layers}
\end{subfigure}

\caption{A comparative analysis of speed and accuracy across various methods.}
\label{fig:inference}
\end{figure}

\textbf{Inference time.}  In Fig. \ref{fig:inference}, we demonstrate the speed-accuracy trade-off between our JCRA and state-of-the-art methods. JCRA surpasses all top-down, bottom-up and end-to-end methods in both speed and accuracy domains. JCRA with ResNet-50 can achieve $78\%$ acceleration of inference speed than previous state-of-the-art bottom-up algorithms.

\subsection{Results on CrowdPose}

The CrowdPose~\cite{li2019crowdpose} dataset is extensive, containing 20,000 images with approximately 80,000 individuals, where each individual was labeled with 14 body joints. It is split into train, validation, and test subsets, containing about 10,000, 2,000, and 8,000 images, respectively. The models were trained on the training and validation subsets to attain optimal performance, and the results were reported on the test set, following the protocol in ~\cite{cheng2020higherhrnet}.

\textbf{Evaluation metric.} The standard average precision based on OKS, which is the same as COCO, is adopted as the evaluation metric. The CrowdPose dataset is divided into three crowding levels: easy, medium, and hard. We report $AP^{kp}$, $AP_{50}^{kp}$, $AP_{75}^{kp}$, $AP_{E}^{kp}$, $AP_{M}^{kp}$, and $AP_{H}^{kp}$ metrics for images in the easy, medium, and hard categories.

\textbf{Test set results.} The results of our approach, compared to other state-of-the-art methods on the test set, are displayed in Table ~\ref{tbk:results}. In contrast to top-down methods that lose their superiority in crowded scenes, our approach demonstrates robustness and achieves a 71.9 AP score without the need for flipping tests, surpassing the most recent two-stage end-to-end method, PETR ~\cite{petr}. Our JCRA is a fully one-stage end-to-end method without a refinement decoder block, which enhances its flexibility and adaptability for estimating human poses in crowded scenes.

\begin{table}[]
\scriptsize
\centering
\begin{tabularx}{0.52\textwidth}{c|c c c|c c c}
\hline
Method&$AP^{kp}$&$AP_{50}^{kp}$&$AP_{75}^{kp}$&$AP_{E}^{kp}$&$AP_{M}^{kp}$&$AP_{H}^{kp}$\\
\hline
\multicolumn{7}{c}{\textbf{Top-down methods}}\\
\hline
Mask R-CNN~\cite{mask}&57.2 &83.5 &60.3 &69.4 &57.9 &45.8\\
SimpleBaseline ~\cite{BinXiao2018SimpleBF}&60.8 &81.4& 65.7 &71.4 &61.2 &51.2\\
SPPE ~\cite{li2019crowdpose}&66.0& 84.2& 71.5 &75.5 &66.3& 57.4\\
\hline
\multicolumn{7}{c}{\textbf{Bottom-up methods}}\\
\hline
HrHRNet~\cite{cheng2020higherhrnet}&65.9 &86.4 &70.6& 73.3 &66.5 &57.9\\
DEKR†~\cite{dekp}&67.3& 86.4& 72.2& 74.6& 68.1& 58.7\\
SWAHR†~\cite{luo2021rethinking}&71.6& 88.5 &77.6 &\textbf{78.9} &72.4 &63.0\\
\hline
\multicolumn{7}{c}{\textbf{Two-stage end-to-end methods}}\\
\hline
PETR~\cite{petr}& 71.6& 90.4& 78.3& 77.3& 72.0& 65.8 \\

\hline
\multicolumn{7}{c}{\textbf{Fully one-stage end-to-end methods}}\\
\hline
JCRA(Ours)~& \textbf{71.9} & \textbf{91.3} & \textbf{78.9} & \textbf{77.8} & \textbf{72.5}& 65.3\\
\hline
\end{tabularx}
\caption{Comparisons with state-of-the-art methods on CrowdPose test dataset. Superscripts E, M, H of AP stand for easy, medium and hard images, respectively. † denotes flipping test.}
\label{tbk:results}
\end{table}

%% file: conclusion.tex
\section{Conclusion and Future Works}
In this work, we present a novel transformer-based network called JCRA, which predicts keypoint results directly. The algorithm is simple, effective, and can be easily designed for real-time industrial multi-person applications, achieving state-of-the-art results on the challenging COCO benchmark. Our experiments validate the ability of JCRA to estimate body poses in high quality and real-time. Further work is needed to improve the accuracy of keypoints for $AP_{M}^{kp}$. Although JCRA achieves similar results for $AP_{L}^{kp}$ compared to top-down methods, improving the accuracy of keypoints for $AP_{M}^{kp}$ will enable JCRA to perform as well as the top-down methods. Additionally, JCRA can be naturally extended by adding an object detection head on top of the decoder outputs, similar to Mask R-CNN's direct addition of a keypoint detection branch from Faster R-CNN. By doing so, the outputs of JCRA will include both keypoints and bounding boxes, allowing the information of bounding boxes to enhance the accuracy of keypoints.